\documentclass[a4paper,twoside]{article}

\usepackage{epsfig}
\usepackage{subcaption}
\usepackage{calc}
\usepackage{amssymb}
\usepackage{amstext}
\usepackage{amsmath}
\usepackage{amsthm}
\usepackage{multicol}
\usepackage{pslatex}
\usepackage{apalike}
\usepackage{algorithm2e}
\usepackage{url}
\usepackage[bottom]{footmisc}
\usepackage{SCITEPRESS}     
\usepackage{listings}
\usepackage{caption} 
\usepackage{tabularx}
\usepackage{tcolorbox} 
\usepackage{courier} 
\usepackage{float} 
\usepackage{hyperref}      

\makeatletter
\newcommand{\confnote}{%
  \begingroup
    \renewcommand\thefootnote{}

    \let\orig@makefntext\@makefntext
    \renewcommand\@makefntext[1]{\parindent0pt\noindent##1}

    \footnotetext{%
      \footnotesize                           
      \noindent\rule{\columnwidth}{0.4pt}\par\vspace{0pt}
      This is the accepted version of the paper presented at the
      \textbf{17th International Conference on Agents and Artificial Intelligence}
      (ICAART 2025), Porto, Portugal.
      Available at: \url{https://doi.org/10.5220/0013321900003890}%
    }%

    \let\@makefntext\orig@makefntext
    \addtocounter{footnote}{-1}%
  \endgroup
}
\makeatother

\tcbset{
  mylistingstyle/.style={
    colback=white, 
    colframe=black, 
    boxrule=0.5pt, 
  },
}

\begin{document}

\title{TRIZ Agents: A Multi-Agent LLM Approach for \\ TRIZ-Based Innovation}

\author{\authorname{Kamil Szczepanik\sup{1}\orcidAuthor{0009-0009-6807-2426}, Jarosław A. Chudziak\sup{1}\orcidAuthor{0000-0003-4534-8652}}
\affiliation{\sup{1}The Institute of Computer Science, Warsaw University of Technology}
\email{\{kamil.szczepanik.stud, jaroslaw.chudziak\}@pw.edu.pl}
}



\abstract{TRIZ, the Theory of Inventive Problem Solving, is a structured, knowledge-based framework for innovation and abstracting problems to find inventive solutions. However, its application is often limited by the complexity and deep interdisciplinary knowledge required. Advancements in Large Language Models (LLMs) have revealed new possibilities for automating parts of this process. While previous studies have explored single LLMs in TRIZ applications, this paper introduces a multi-agent approach. We propose an LLM-based multi-agent system, called TRIZ agents, each with specialized capabilities and tool access, collaboratively solving inventive problems based on the TRIZ methodology. This multi-agent system leverages agents with various domain expertise to efficiently navigate TRIZ steps. The aim is to model and simulate an inventive process with language agents. We assess the effectiveness of this team of agents in addressing complex innovation challenges based on a selected case study in engineering. We demonstrate the potential of agent collaboration to produce diverse, inventive solutions. This research contributes to the future of AI-driven innovation, showcasing the advantages of decentralized problem-solving in complex ideation tasks.}

\onecolumn \maketitle \normalsize \setcounter{footnote}{0} \vfill

\confnote   

\section{INTRODUCTION}
\label{sec:introduction}

Problem-solving is a fundamental subject of every innovation process. It is much more than finding quick fixes or solutions to occurring obstacles. Innovation problem-solving requires substantial knowledge and unconventional thinking. However, it is strongly advised to follow methodologies designed to succeed in this process. Structured methodologies for problem-solving in innovation, like TRIZ (Theory of Inventive Problem Solving) \cite{orloff2006triz}, Design Thinking \cite{brown2008design}, Lean Innovation \cite{sehested2010lean}, and more, were created. 

Researchers have preferred or disliked methods; some work only for specific problems or become hard to implement. The innovation process is a challenging task, and every tool that tries to harness it is merely a rough guide for innovators during the process. Nevertheless, these methodologies are the best-known way to cope with it.

TRIZ has been around for decades, proving its efficiency and usefulness as a problem-solving methodology. It was developed after analyzing around 200,000 patents \cite{loh2006automatic} to create a structured framework for engineers and innovators to solve innovation problems. Over the years, many improvements have been proposed; however, the general idea has stayed the same. In addition, complementary tools and group work techniques have been developed to assist teams during the TRIZ process, such as brainstorming \cite{mahto2013concepts}.

Although the TRIZ methodology has been successfully helping innovators solve problems, it has limitations that still need to be addressed. The biggest one is perhaps the need for experts. The only way to solve complex innovation challenges is for innovators to have extensive knowledge and experience in the relevant field \cite{czinki2016solving} \cite{ilevbare2013review}. It takes many years to become an expert in one domain, and sometimes, expertise in multiple domains is required to solve complex innovation problems. People working in an innovation team must think creatively and possess broad knowledge. There is a need for interdisciplinary knowledge when creating a TRIZ team. There have been proposed concepts supporting knowledge discovery \cite{5352735}; however, with the advent of LLMs, new possibilities have arisen.

This study simulates a realistic team tasked with solving innovation problems using the TRIZ methodology. It aims to model agents' subtasks, goals, and abilities for collaboratively achieving a theoretical solution proposal. Choosing the TRIZ methodology narrows the scope of actions while guiding the team in a structured and systematic way.

The primary purpose is to showcase the abilities of language agents in collaborative teamwork. This study proposes a multi-agent system in which LLM agents simulate a team's workflow to solve innovation problems using the TRIZ methodology. First, we review previous works on this topic and introduce the reader to the TRIZ methodology. Next, we describe the workflow design, defining key assumptions and the system's scope. The architecture and implementation details are discussed, outlining agent orchestration, agents' abilities, and tools. The subject of the system's work is a case study of the TRIZ application for the improvement design of a gantry crane \cite{luing2024application}, which precisely documents each step of the researchers' workflow. Lastly, the system results are assessed and discussed.

The experiments follow the ideation process of agents solving the case study problem, with each step's outcome assessed against the original study. The focus is not on comparing agent orchestration architectures or prompt strategies.

Exploring LLM-based multi-agent systems may significantly enhance problem-solving tasks requiring creative thinking, external resources, and broad knowledge. Agent collaboration methods are likely to become more investigated because of their potential to automate processes and optimize business operations.

The paper is structured as follows. Section 2 describes the background and related work. Section 3 presents our TRIZ Agents framework. Section 4 describes our case study. Section 5 discusses the system results and Section 6 summarizes our conclusions.

\section{BACKGROUND}

\subsection{Related Work}
Previous research has explored using LLMs to enhance the innovation process with TRIZ methodology. However, the focus was more on creating workflows for such models \cite{chen2024improvinglargemodelssmall}, comparing prompting strategies and LLM models \cite{jiang2024autotrizartificialideationtriz}. Previous works include building workflows that strictly follow TRIZ steps and solve those steps with refined, prompt engineering strategies. Results showed a promising area of LLM usage - innovative problem-solving.

Studies have also suggested that agentic systems based on LLMs are significantly effective \cite{wu2023autogenenablingnextgenllm}. With agentic GPT-3.5, it was possible to achieve results better than the foundation GPT-4 model, a next-generation model \cite{hu2024automateddesignagenticsystems}. This sparks a need to explore this approach. 

Over the past few years, researchers and providers have developed numerous models. OpenAI's GPT model is the most widely used, with many integrations, which is why we decided to utilize it for our experiments. Using other available LLMs, such as Gemini, Llama, or Claude, would be equally justified. This research does not focus on comparing LLMs.

Recent advancements in artificial intelligence, especially natural language processing, have significantly increased possible AI applications. The emergence of Large Language Models and intensive research in this area has shown that many new domains can be supported with AI. Models, like OpenAI's GPT-3, have extensive knowledge about the world \cite{atox2024evaluating}. They have shown impressive performance in passing tests concerning intelligence and problem-solving abilities \cite{orru2023human}. Surprisingly, LLMs have also demonstrated remarkable capabilities in creative tasks, such as image generation \cite{oppenlaender2022creativity}, writing \cite{chakrabarty2024creativity}, and even music generation \cite{yuan2024chatmusicianunderstandinggeneratingmusic}. These results suggest using LLMs to solve problems that require creative thinking, such as designing innovations. Since current research suggests that it is best to guide LLMs and carefully structure their tasks \cite{minaee2024largelanguagemodelssurvey}, the TRIZ methodology seems perfect, with clear, progressive steps and guidelines.

\begin{figure}
  \centering
   \includegraphics[width=0.5\columnwidth]{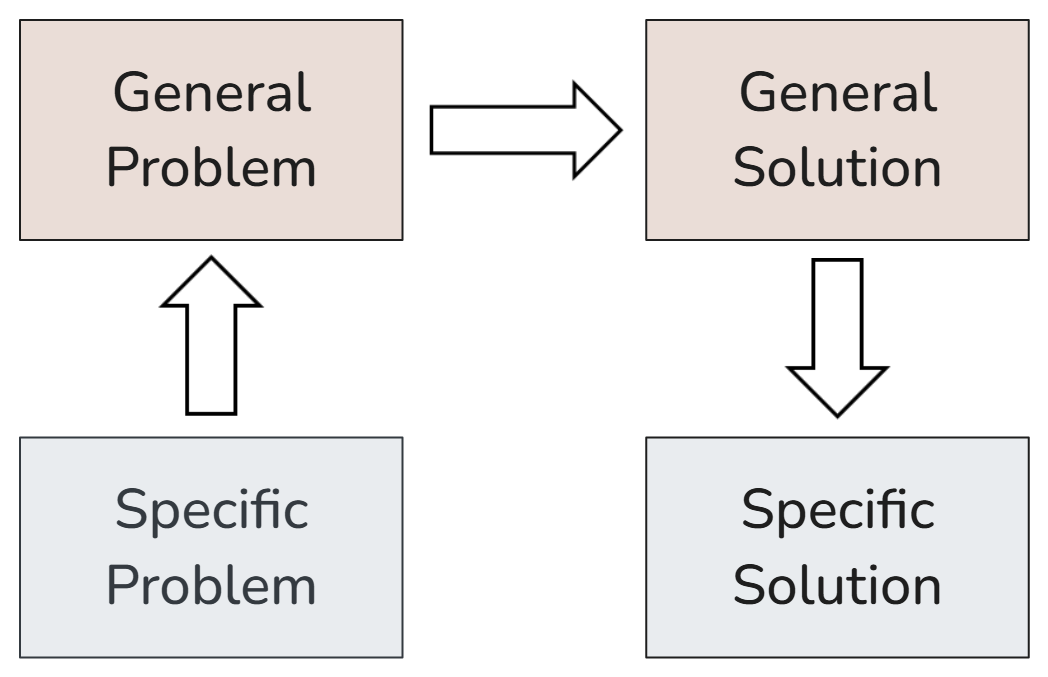}
  \caption{TRIZ Problem Solving Methodology.}
  \label{fig:triz_stages}
 \end{figure}

\subsection{TRIZ Methodology}

TRIZ is a well-known and broadly used method of solving innovation problems. It was proposed by Genrich Altshuller in the 1960s \cite{arciszewski2016inventive}. It introduced a systematic approach to innovative problem-solving. A critical insight from TRIZ creators is that problems and solutions reoccur across domains of industries and sciences. The core idea behind it is an assumption that there must be something in common, some pattern, in solving problems described in thousands of patents. Four stages of TRIZ are shown in Figure \ref{fig:triz_stages}. This analysis allowed researchers to define a step-by-step method for overcoming innovation obstacles by abstracting them into generalized patterns of invention. This allowed many innovators to overcome seemingly impossible challenges without the need for unfavorable compromise. Although TRIZ has primarily involved mechanical problems, it was effectively transferred to other domains, making it a universal method for innovative problem-solving. 

 \begin{figure}[!b]
  \centering
   \includegraphics[width=0.8\columnwidth]{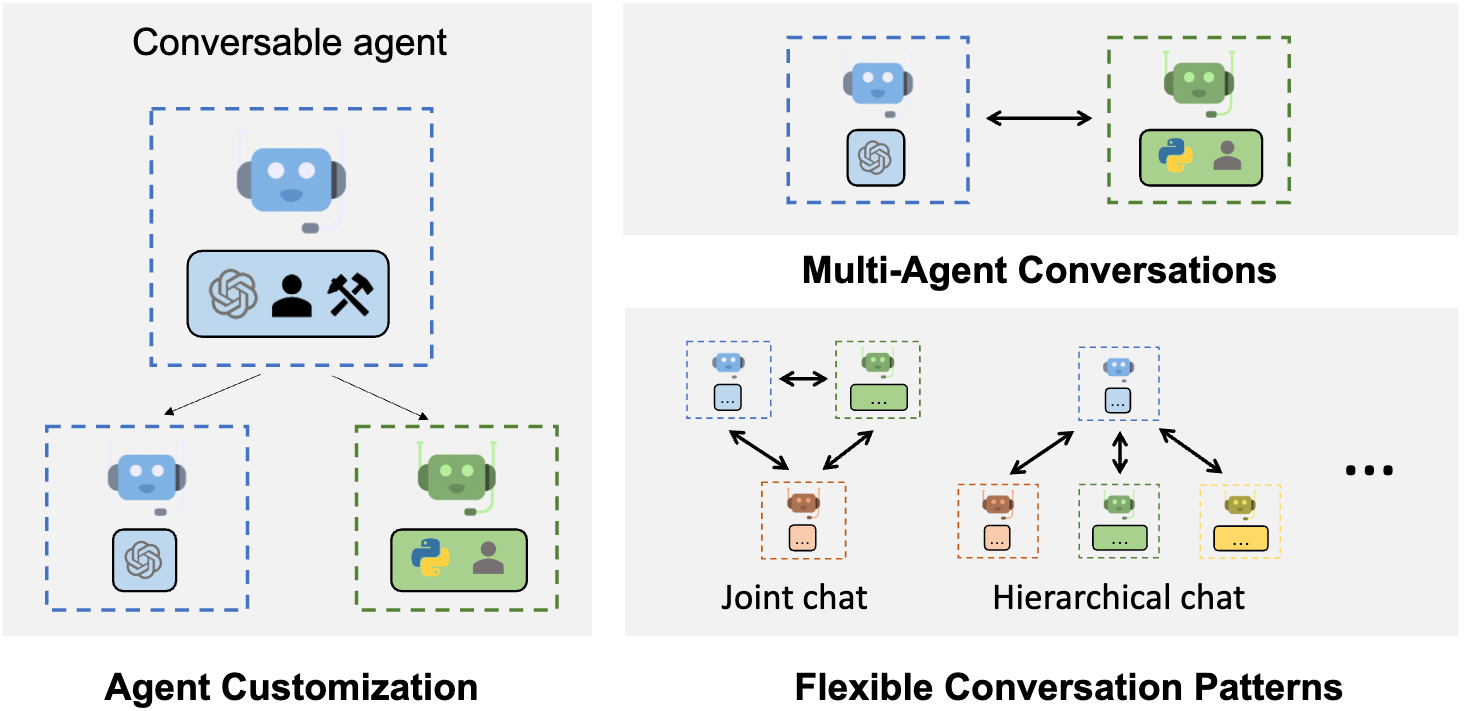}
  \caption{Agents collaboration methods \cite{wu2023autogenenablingnextgenllm}.}
  \label{fig:autogen}
 \end{figure}

\begin{figure*}[htbp]
    \centering
    \includegraphics[width=0.9\textwidth]{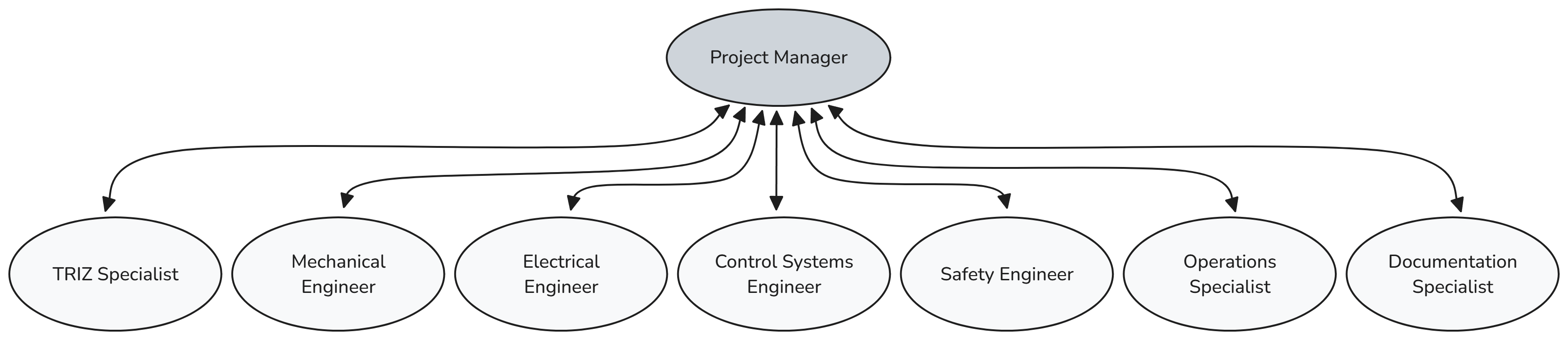}
    \caption{TRIZ Agents diagram, showing connections between key agents.}
    \label{fig:wide}
\end{figure*}

\subsubsection{TRIZ Core Principles}

The core principle of TRIZ is resolving contradictions identified in the problem matter, where often, improving one feature implies worsening another. The TRIZ methodology provides tools to support the resolution of these contradictions. Two primary tools are the 40 Inventive Principles and the Contradiction Matrix.

The TRIZ process starts with a problem analysis, resulting in the identification of TRIZ parameters. According to TRIZ methodology, 39 parameters - such as Speed, Force, Temperature, etc. - represent system characteristics. By identifying the parameters relevant to the specific problem and their interdependencies, the problem can be generalized and structured within TRIZ. 

The Contradiction Matrix is a tool for finding Innovation Principles by providing two TRIZ parameters: the improving feature and the worsening feature. The matrix contains information on which contradictions can be solved by which principles. The study of thousands of patterns enabled the structuring of parameters and the creation of this tool. It is worth noting that the contradiction matrix consists of empty cells, meaning there are no inventive principles for some contradictions.

The 40 Inventive Principles define ideas and approaches for resolving contradictions. They serve as instructions or guidelines for overcoming given conditions. They provide solutions to abstract problems, which can be transformed into practical solutions in the real world.

 \begin{figure}[!b]
  \centering
   \includegraphics[width=0.9\columnwidth]{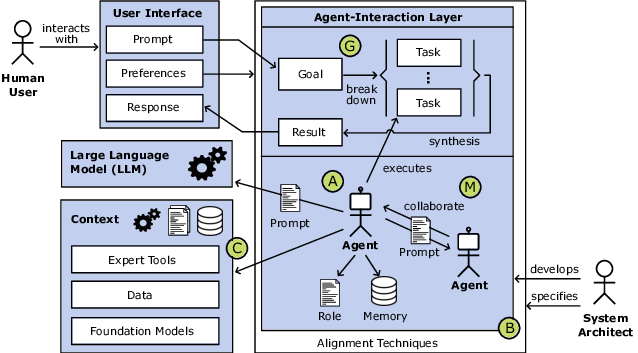}
  \caption{Architecture of LLM-based multi-agent system enhanced with tools and data sources \cite{handler2023taxonomy}.}
  \label{fig:multiagentsystem}
 \end{figure}
 
\subsubsection{TRIZ Complexity and Knowledge Demands}
TRIZ application follows clear guidelines; however, to abstract problems effectively, identify contradictions, and use TRIZ tools, it still requires deep interdisciplinary knowledge of innovators \cite{moehrle2005triz}. This makes it challenging to apply for non-experts, and the final result is highly dependent on innovators' expertise due to the high complexity of the problems. 
In addition, following TRIZ steps like analyzing system parameters, identifying contradictions, or scrutiny of inventive principles is very labor-intensive and time-consuming \cite{nassar2016introduction} \cite{cascini2004plastics}. 
For this reason, resulting in limited accessibility, TRIZ has been adopted only in specialized innovation environments like R\&D centers or academic sites. Given its effectiveness, it would be desirable to implement it in a broader range of environments, especially those without resources or access to experts.

\subsection{Multi-agent Systems and Language Agents}

Multi-Agent Systems (MAS), a subdomain of Distributed Artificial Intelligence (DAI), is a system that consists of multiple autonomous entities known as agents \cite{dorri2018multi}. Each agent, put together in a shared environment, can operate independently but communicate and collaborate with others or interact with the system to solve more complex tasks \cite{van2008handbook}. Agents alone cannot perform complex tasks; however, after decomposing them into smaller subproblems, specialized agents can solve them and, therefore, solve the primary task together. MAS are valued for flexibility and adaptability, especially in distributed intelligence and decision-making \cite{serugendo2005self}. Figure \ref{fig:multiagentsystem} presents an exemplary architecture of an LLM-based system, where tasks are distributed to multiple agents equipped with context sources like tools or databases. Those systems seem to have a significant amount of use cases because they, in a way, simulate the work of human workers. 

As opposed to traditional MAS that use formal agent communication language protocols such as FIPA ACL \cite{kone2000state}, our proposed approach utilizes natural language prompts. Moreover, agents communicate with one another by natural language. While FIPA ACL provides a good, standardized, and rigorous way for agent interactions, LLM-based communication offers flexibility and adaptability for complex collaborative tasks.
 
Multi-agent systems based on LLMs are currently being researched extensively. Their core idea is to handle complex problems through collaboration and language abilities \cite{NEURIPS2023_65a39213}. 
A notable example of a multi-agent system is presented in the study \cite{park2023generativeagentsinteractivesimulacra}, where agents demonstrate human-like behavior within a shared interactive environment. This research underscores the potential of LLM-based multi-agent systems for simulating complex human communities.
Studies have explored various implementations of teams, such as a software engineering team \cite{chudziak_cinkusz_llm} \cite{electronics14010087} and a multi-agent system designed for big data analysis in financial markets \cite{app142411897}. Other research focuses on enhancing agents' logical reasoning capabilities and memory functions within LLM-based multi-agent systems \cite{synergymas}.

\subsubsection{Agent Orchestration}

Along with research about the application of LLMs in multi-agent systems, a new subdomain of those studies has emerged called Agent Orchestration. It explores how different agent architectures work and produce results. In the past, there was no need for that because most multi-agent systems were deterministic microservices, and in general, the result of given input could be known before running the system. LLM agents introduced new areas of research because of their stochasticity, flexibility, and ability to adapt or remember abstract information \cite{cheng2024exploringlargelanguagemodel}. This promising field of study involves designing and evaluating agent orchestration architectures. It includes investigating various ways agents interact, such as supervised groups, equal collaboration, and nested collaborating teams, as shown in Figure \ref{fig:autogen}. Agent orchestration techniques optimize the abilities of each agent and, therefore, synergetically obtain the best results. 

Another related research field is prompt engineering. This domain focuses on optimizing prompts to achieve the best LLM outputs. Other studies demonstrate that employing specific prompting strategies — such as reasoning and acting (ReAct) \cite{yao2023reactsynergizingreasoningacting} or Chain-of-Thought Prompting \cite{wei2023chainofthoughtpromptingelicitsreasoning} — can significantly enhance the quality of generated outcomes. 
The way in which roles and tasks are described to an LLM agent is essential, and the outcome strongly depends on it \cite{zamfirescu2023johnny}. It has to be clear what the structure of the agent's output should be. This study focuses on basic prompt strategy, which consists of tasks, roles, and contexts for the agent.

\subsubsection{Agent Tools}
What we know about agentic systems is that you can never fully predict the outcome of an LLM’s actions. While LLMs are pretty proficient at planning and decision-making, a default LLM model cannot perform actions or directly interact with the real world. Tools enable these interactions by allowing LLMs to generate function call outputs with specified arguments. This mechanism can equip models with tools such as web search, database queries, file reading and writing, and more. Our goal is to provide agents with tools to take action, apply grounding, and interact with the system.

A known characteristic of LLMs is their tendency to hallucinate, which comes from the fundamental mathematical and logical structure of these models \cite{banerjee2024llmshallucinateneedlive}. 
Although LLM knowledge is broad and often accurate, it is not advisable to assume that every generated text is truthful. This is especially important in our case, as we want agents to reason based on fact-checked information. One way to minimize the generation of incorrect knowledge is by using a Retrieval-Augmented Generation (RAG) mechanism 
\cite{béchard2024reducinghallucinationstructuredoutputs}. 
By incorporating RAG into a tool, agents can check necessary knowledge by retrieving it from predefined and reliable sources. 

A simple web search is another helpful tool that helps ground the models. It allows browsing the internet with queries about the topic. Eventually, the agent gets valuable information, and based on it, it generates a response.

\subsection{Role of LLMs in Problem-Solving and Innovation}

Artificial intelligence has been applied to various and numerous tasks, and TRIZ is one of them. Those improvements include estimation of ideality and level of invention \cite{adams2009computer}, or classification and analyzing engineering patents \cite{hall2022triz}. Recently, LLMs have opened a new set of possible TRIZ enhancements. In recent works, researchers have been exploring ways of using LLMs' extensive broad knowledge and substantial reasoning abilities to create systems supporting innovators in the ideation process or entirely replace them \cite{chen2024trizgptllmaugmentedmethodproblemsolving}\cite{jiang2024autotrizartificialideationtriz}. Those researchers have uncovered the potential of LLMs in applications requiring creative thinking.
The limitations of these methods become apparent with larger and more complex tasks. Given that LLMs perform more effectively on smaller subtasks \cite{chen2024improvinglargemodelssmall}, exploring Multi-Agent Systems has emerged as a natural solution for handling such complexity. 
Such systems could be flexible and adaptable to the given problem, enabling interdisciplinary problem-solving capabilities.

\section{TRIZ AGENTS}

\begin{figure}
    \centering
    \includegraphics[width=5.5cm]{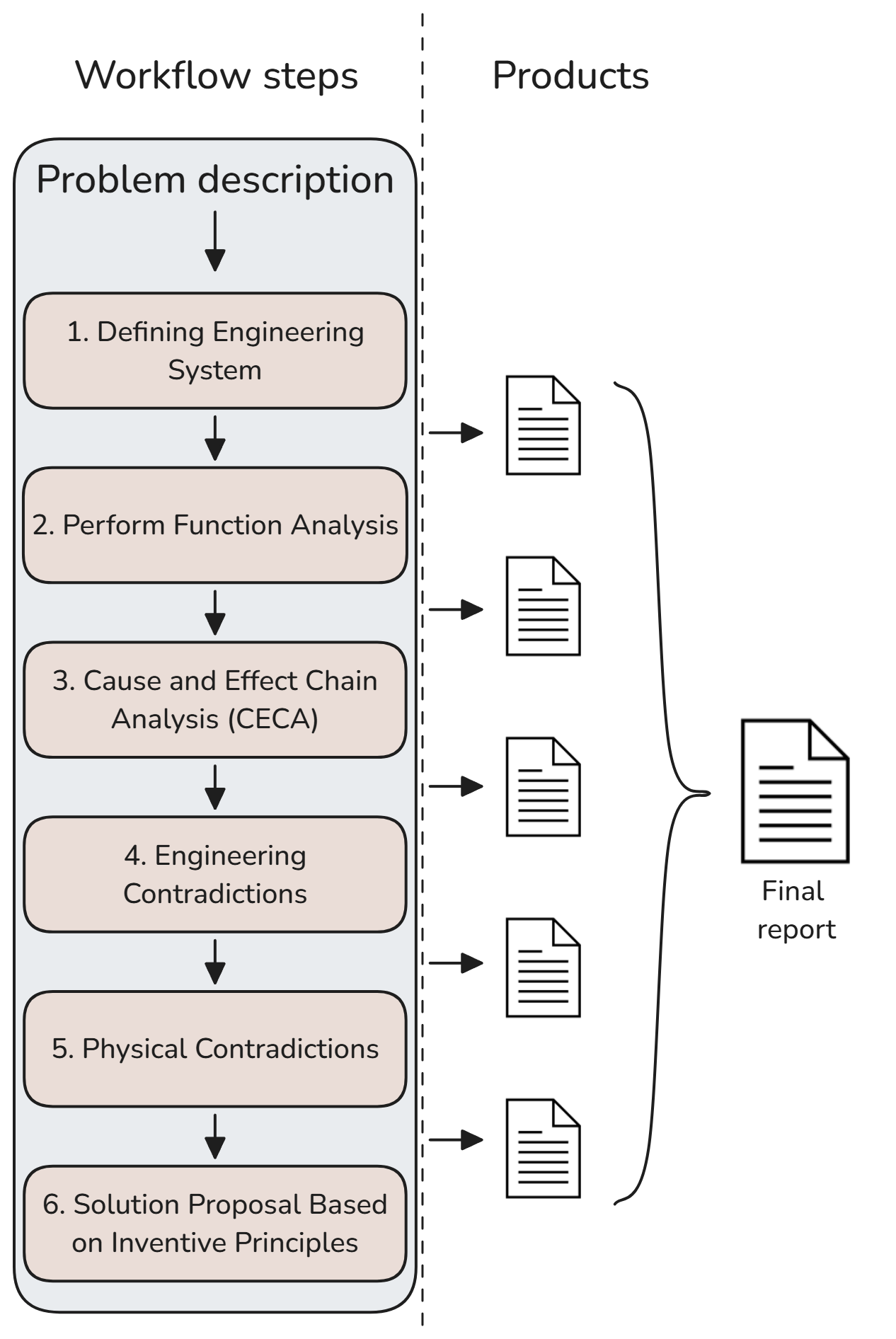}
    \caption{TRIZ Agents workflow steps and products.}
    \label{fig:workflow}
\end{figure}

\subsection{Workflow}

\begin{figure}[!b]
  \centering
   \includegraphics[width=0.8\columnwidth]{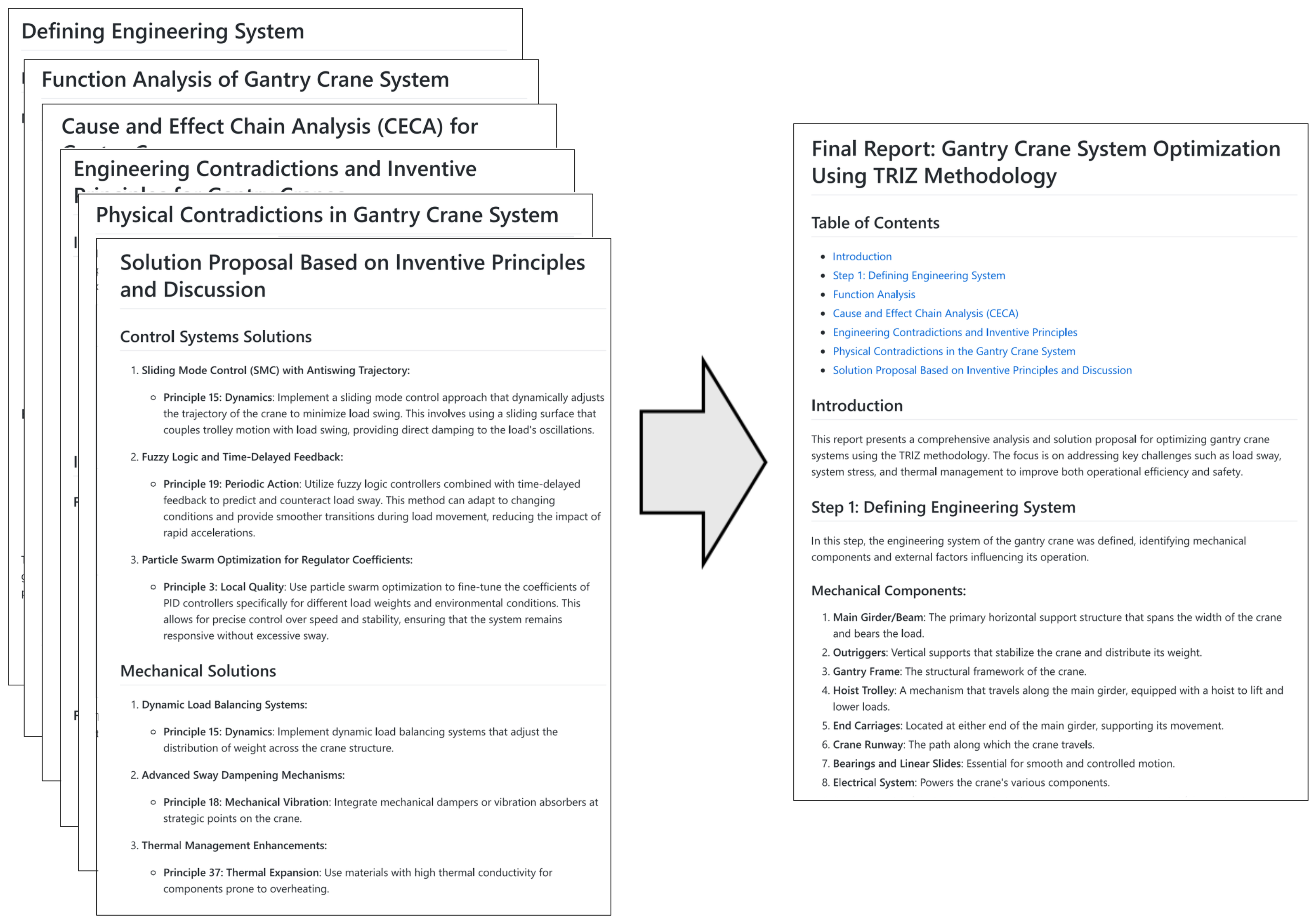}
  \caption{TRIZ Agents products.}
  \label{fig:products}
 \end{figure}

As said before, the general idea simulates an innovation team's ideation process. The team is provided with a problem description and, in return, must propose one or more solutions. The team works based on the steps indicated in the case study paper. Each step is done one by one, and a special agent documents each step. Documentation of the previous step is provided to the team. One way to think about it is that each step is a separate meeting where team members solve a problem. At the next meeting, team members don't remember what they said last time, but they have documentation and the main findings of the previous step. Based on that, they continue with the following step. The process repeats until the last step is achieved. After that, another documentation agent compiles all documentation into one concise report (Figure \ref{fig:products}). 

The ideal result would be that the team produces the exact solutions with the same intermediate steps, and the final report's content is the same as the case study's content. Each step will be compared and evaluated with the original case study. All steps are presented in Figure \ref{fig:workflow} for better workflow clarity. Documenting the team's progress at each step helps to track the progress, but it also allows us to compare results to the original case study's corresponding parts.

\subsubsection{Implementation}

System implementation is based on LangGraph \cite{langgraph_2023}, a framework designed to build applications that integrate LLMs. They provide precise agent orchestration and control over the graph of agents. Each agent was implemented as a node in a graph, where edges are connections between them - conditional or not. The agents' tools are separate nodes as well. This framework allows for more structured LLM text generation by using prompt templates while calling LLM models. Messages of each step are the primary context for LLM chat models to generate answers. It is worth mentioning that in LangChain, messages have classes, including Human or AI. In the system, messages generated by models are converted to human type so that the next model generates outputs as if it were conversing with a human. 

For the LLM model, GPT-4o was selected due to its high performance and relatively fast response time. It supports function calls, which is necessary for tool integration. Outputs of LLMs rely on the model temperature parameter. In this study, this parameter was arbitrarily set to 0.5 since it is a good compromise between reasonable and creative answers.

\begin{figure}
    \centering
    \includegraphics[width=0.55\columnwidth]{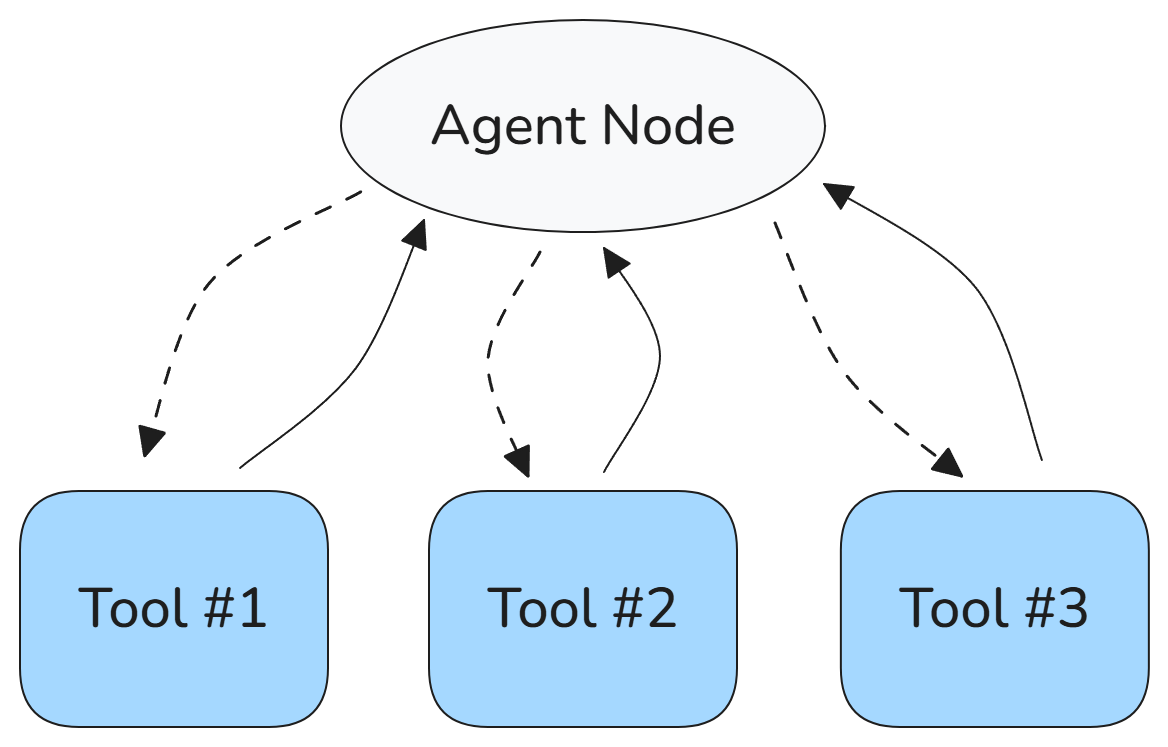}
    \caption{Agent node with tool nodes.}
    \label{fig:agent_nodes_tools}
\end{figure}

\subsection{Agent Orchestration Architecture}

For agent orchestration, a supervised team was chosen. In the experiment, we call this agent a \texttt{Project Manager}. This agent is responsible for following the workflow, distributing tasks, and telling who should act next. This agent, an orchestrator, manages the team and makes sure it goes in the right direction. The simple way to think about it is an actual group leader in a real-life scenario, but its capabilities and responsibilities are a bit narrow for the LLM efficacy. Provided with workflow wrought on the case study paper, \texttt{Project Manager} keeps track of the project and steers the conversation so that the team solves tasks one by one. Agents engage in the conversation once they are asked to by \texttt{Project Manager}. They perform necessary actions, like tool usage, and provide an appropriate answer. After that, attention is brought back to \texttt{Project Manager}, who receives the message and continues with management.

\subsubsection{Team Composition}
There are the following agents in the team: \texttt{Mechanical Engineer}, \texttt{Electrical Engineer}, \texttt{Control Systems Engineer}, \texttt{Safety Engineer}, \texttt{TRIZ Specialist}, \texttt{Operations Specialist}, \texttt{Documentation Specialist} and supervisor \texttt{Project Manager}. Each agent's role was defined based on case study analysis, which allowed to specify what members the team would consist of.

\subsubsection{Agent Profiling Prompts}

 Each agent has its own prompt specifying their persona. It gives information about their field of expertise, tools, and what their output should look like. Each agent needs a profiling prompt, which will define the persona behind it. Key elements that need such prompts are:
\begin{itemize}
    \item \textbf{Name}: Name of the agent. Necessary for finding oneself in conversation.
    \item \textbf{Role}: Basic description of the agent's place in the team and its background for character development.
    \item \textbf{Tasks and Reponsibilities}: Clearly states their tasks and responsibilities so that LLM's focus is primarily on them.
    \item \textbf{Context}: Gives the agent an idea of what is the current stage of the process. It can be messages of the conversation or a summary of previous messages.
\end{itemize}

\begin{figure}
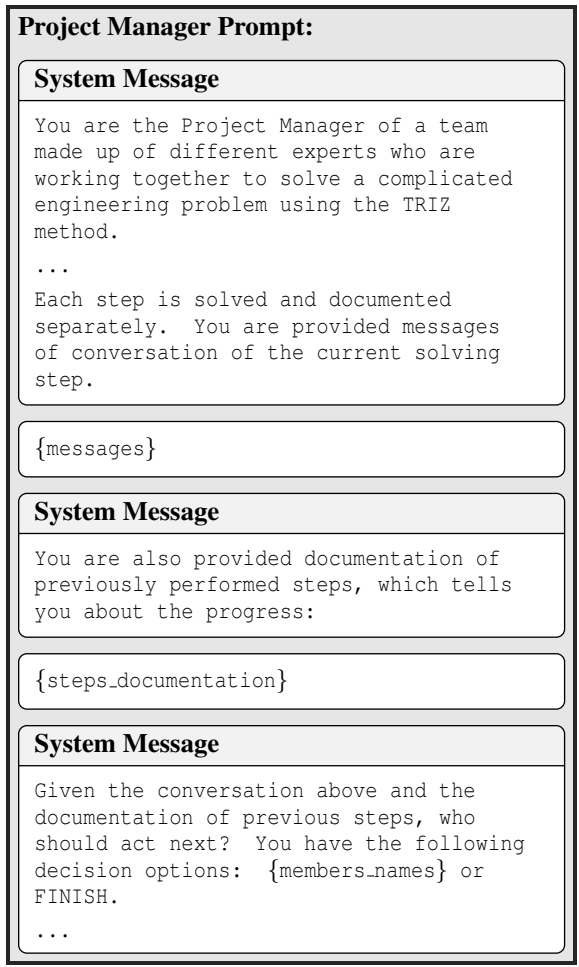

    \centering
\noindent
\begin{minipage}{\columnwidth}
\begin{tcolorbox}[colback=gray!20, colframe=black!85, boxrule=0.5mm, sharp corners,
  left=0.5mm, right=0.5mm, top=0.5mm, bottom=0.5mm, boxsep=0.5mm]

\noindent\textbf{Project Manager Prompt:}
\begin{tcolorbox}[mylistingstyle,
  title=System Message, 
  coltitle=black,
  fonttitle=\bfseries,
  fontupper=\small, 
  colbacktitle=gray!10,
  colframe=black, 
  colback=white, 
  left=1mm, right=1mm, top=1mm, bottom=1mm, boxsep=1mm
]
\texttt{You are the Project Manager of a team made up of different experts who are working together to solve a complicated engineering problem using the TRIZ method.}

\vspace{0.1cm}
\texttt{...}
\vspace{0.1cm}

\texttt{Each step is solved and documented separately. You are provided messages of conversation of the current solving step.}
\end{tcolorbox}
\begin{tcolorbox}[mylistingstyle,colframe=black, colback=white, fonttitle=\bfseries, fontupper=\small, left=1mm, right=1mm, top=1mm, bottom=1mm, boxsep=1mm]
\texttt{\{messages\}}
\end{tcolorbox}

\begin{tcolorbox}[mylistingstyle,title=System Message, coltitle=black, fonttitle=\bfseries, fontupper=\small, colbacktitle=gray!10, colframe=black, colback=white,
  left=1mm, right=1mm, top=1mm, bottom=1mm, boxsep=1mm
]
\texttt{You are also provided documentation of previously performed steps, which tells you about the progress:}
\end{tcolorbox}
\begin{tcolorbox}[mylistingstyle,colframe=black, colback=white,fonttitle=\bfseries, fontupper=\small, left=1mm, right=1mm, top=1mm, bottom=1mm, boxsep=1mm]
\texttt{\{steps\_documentation\}}
\end{tcolorbox}

\begin{tcolorbox}[mylistingstyle,title=System Message, coltitle=black, fonttitle=\bfseries, fontupper=\small,colbacktitle=gray!10, colframe=black, colback=white,
  left=1mm, right=1mm, top=1mm, bottom=1mm, boxsep=1mm
]
\texttt{Given the conversation above and the documentation of previous steps, who should act next? You have the following decision options: \{members\_names\} or FINISH.}

\vspace{0.1cm}
\texttt{...}
\end{tcolorbox}
\end{tcolorbox}
\caption{Example of \texttt{Project Manager} prompt template. Elements in \{\} brackets are inputs of prompt templates.}
\label{fig:manager_prompt}
\end{minipage}
\end{figure}

In addition, it is advisable to add some specific instructions in the prompt that are critical for correct system operation or just essential in the agent's role. The prompt template of \texttt{Project Manager} agent is presented in Figure \ref{fig:manager_prompt}.

As it is shown, prompts are composed of system messages and placeholders, such as messages in the conversation or documentation on previous steps. It is necessary to provide all inputs to the template in order to run the model.

\subsubsection{Tools of TRIZ Agents}

In the real world, team members can use various tools and knowledge bases, which is why we will provide agents with the same. All agents except \texttt{Project Manager} are equipped with tools. In the LangGraph framework, agent nodes have conditional connections with tool nodes since agents may or may not want to use them. Contrarily, the edge from the tool node to the agent node is non-conditional, so the right agent gets the tool's output. Figure \ref{fig:agent_nodes_tools} presents a worker agent node with tool nodes.

The composition of the toolset for an agent depends on their role in the team. All sets include a web search tool, which is their elemental way of acquiring knowledge. Since \texttt{TRIZ Specialist} is an important team member, this agent is additionally equipped with TRIZ-oriented tools such as:
\begin{itemize}
    \item TRIZ Features Tool: Return list of 39 TRIZ parameters. 
    \item Contradiction Matrix Tool: Returns worth applying inventive principles based on improving and worsening features of the system.
    \item TRIZ Inventive Principles Tool: Tool returns more details about given inventive principles.
    \item TRIZ RAG Tool: Returns answers to agent's queries based on TRIZ source materials, such as books, websites, Wikipedia pages, etc.
\end{itemize}

\section{CASE STUDY}
To model team cooperation and evaluate the results of the proposed agentic system, we conducted a case study about the application of TRIZ for gantry crane improvement based on \cite{luing2024application}. The authors of this paper thoroughly describe each step of their work. They strictly follow the TRIZ methodology. The final outcome of the ideation process consists of textual descriptions of the proposed approaches. During the problem analysis, authors identify system elements, perform CECA analysis, find physical and engineering contradictions, and apply TRIZ principles using the TRIZ Matrix. In addition, authors build graphs for analysis steps, which is again feasible using LLMs. All those steps are a great usage scenario for a team of agents. A team of predefined team members will cooperate and invent a solution for the crane improvement problem.

It is worth mentioning the criteria for the chosen case study. First of all, the problem description is sufficiently characterized. All the intermediate steps and solutions are clearly explained, which is vital during the evaluation of results. Moreover, a case study was published in the year 2024, which means that Chat GPT-4o used in experiments was for sure not trained on this text because the data cutoff for this model is October 2023. This is extremely important since we don't want the model to know the solution - we want the system to invent it.

\section{RESULTS AND DISCUSSION}

The proposed multi-agent system simulates a discussion between team members as intended, resulting in six documents about each workflow step and one final report, which compiles those documents. As an input system message, the system receives a Human type message with textual problems description of the case study (Figure \ref{fig:input_message}). It is copied from the case study article not to omit information or add one.

\begin{figure}[!ht]
    \centering
\begin{tcolorbox}[mylistingstyle,
  title=Input Human Message, 
  coltitle=black,
  fonttitle=\bfseries,
  fontupper=\small, 
  colbacktitle=gray!10,
  colframe=black, 
  colback=white,
  left=1mm, right=1mm, top=1mm, bottom=1mm, boxsep=1mm
]
\texttt{Solve the following problem: Gantry cranes find extensive application across various industries, employed to move hefty loads and dangerous substances within shipping docks, building sites, steel plants, storage facilities, and similar industrial settings. The crane should move the load fast without causing any unnecessary excessive swing at the final position. Moreover, gantry cranes which always lift excessive load may result sudden stop of the crane. The crane operators' attempt to lift heavier loads at a faster pace has led to recurrent malfunctions, including overheating, and the increased speed has caused excessive swinging or swaying of the lifted load, posing a safety hazard.}
\end{tcolorbox}
\caption{TRIZ Agents input message taken from case study problem description.}
\label{fig:input_message}
\end{figure}

It is essential to note the system's outputs are not deterministic, which comes from the nature of LLMs \cite{Ouyang_2024}. That is why the results described here are generalized descriptions of how the majority of runs went. 

Another observation worth mentioning is the number of iterations in the agent graph and the number of total tokens (prompt and output tokens). By iterations in the graph, we mean the number of times any node was called. This number varied from 60 to 80 calls. Total tokens of one runtime varied from 150,000 to 250,000.

\subsection{Analysis of System's Actions}

\paragraph{Step 1: Defining Engineering System} \texttt{Project Manager} prompts \texttt{Mechanical Engineer} to identify relevant elemnts of gantry crane engineering system. This agent uses a web search tool to find relevant information and, based on that, specifies the engineering system. After that, \texttt{Project Manager} decides to finish this step by documenting findings by \texttt{Document Specialist}.

\paragraph{Step 2: Function Analysis} \texttt{Project Manager} has at disposal the documentation from previous step. Based on that, \texttt{Project Manager} this time asks \texttt{Mechanical Engineer} to prepare the Function Analysis. The agent uses a web search tool to generate an analysis. Next, a \texttt{Control Systems Engineer} is asked to do the same from the control systems engineering perspective. Like the previous agent, it uses a web search tool to generate analysis. Finally, \texttt{Safety Engineer} is asked to evaluate the safety features and potential hazards of the system. After that, \texttt{Project Manager} decides to move on to the next step by documentation of \texttt{Documentation Specialist}.

\begin{figure}
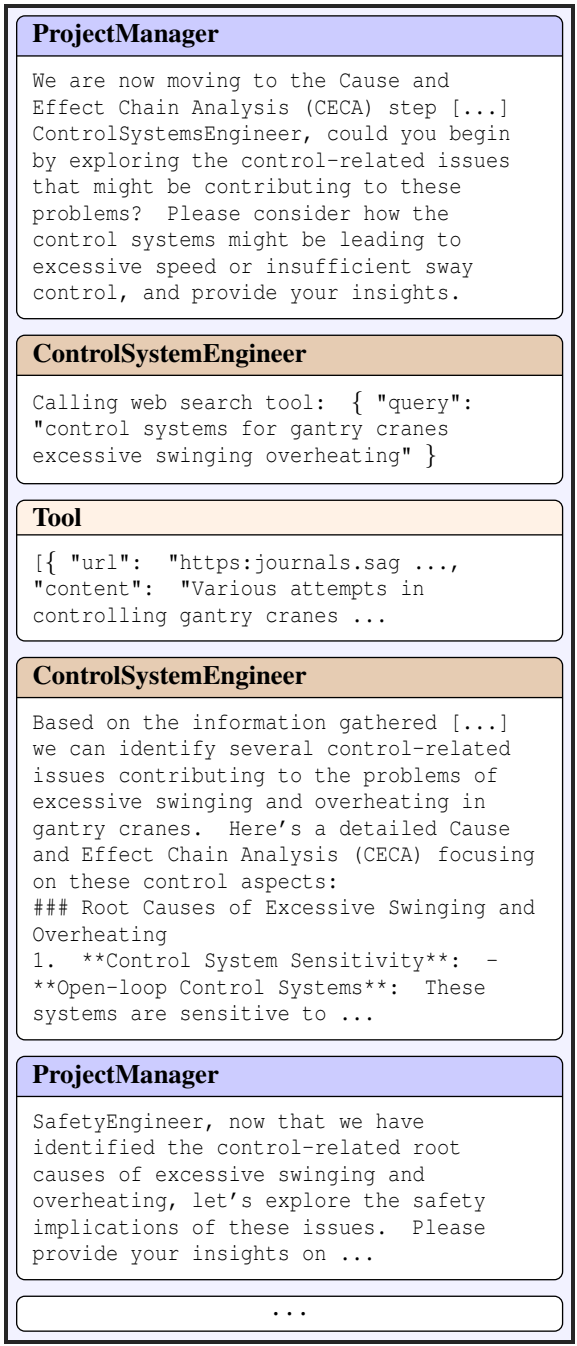

    \centering
\noindent
\begin{minipage}{\columnwidth}
\begin{tcolorbox}[colback=blue!5, colframe=black!85, boxrule=0.5mm, sharp corners,
  left=0.5mm, right=0.5mm, top=0.5mm, bottom=0.5mm, boxsep=0.5mm]

\begin{tcolorbox}[mylistingstyle,
  title=ProjectManager, 
  coltitle=black, fonttitle=\bfseries, fontupper=\small, colbacktitle=blue!20, colframe=black, colback=white,
  left=1mm, right=1mm, top=1mm, bottom=1mm, boxsep=1mm]
\texttt{We are now moving to the Cause and Effect Chain Analysis (CECA) step [...] ControlSystemsEngineer, could you begin by exploring the control-related issues that might be contributing to these problems? Please consider how the control systems might be leading to excessive speed or insufficient sway control, and provide your insights.}
\end{tcolorbox}
\begin{tcolorbox}[mylistingstyle,
  title=ControlSystemEngineer, 
  coltitle=black, fonttitle=\bfseries,fontupper=\small, colbacktitle=brown!40, colframe=black, colback=white,
  left=1mm, right=1mm, top=1mm, bottom=1mm, boxsep=1mm]
\texttt{Calling web search tool:
\{
  "query": "control systems for gantry cranes excessive swinging overheating"
\}
}
\end{tcolorbox}
\begin{tcolorbox}[mylistingstyle,
  title=Tool, 
  coltitle=black, fonttitle=\bfseries, fontupper=\small,colbacktitle=orange!10, colframe=black, colback=white,
  left=1mm, right=1mm, top=1mm, bottom=1mm, boxsep=1mm]
\texttt{[\{
    "url": "https:\/\/journals.sag ...,
    "content": "Various attempts in controlling gantry cranes ...
}
\end{tcolorbox}
\begin{tcolorbox}[mylistingstyle,
  title=ControlSystemEngineer, 
  coltitle=black, fonttitle=\bfseries,fontupper=\small, colbacktitle=brown!40, colframe=black, colback=white,
  left=1mm, right=1mm, top=1mm, bottom=1mm, boxsep=1mm]
\texttt{Based on the information gathered [...] we can identify several control-related issues contributing to the problems of excessive swinging and overheating in gantry cranes. Here's a detailed Cause and Effect Chain Analysis (CECA) focusing on these control aspects:}

\texttt{\#\#\# Root Causes of Excessive Swinging and Overheating}

\texttt{1. **Control System Sensitivity**:
   - **Open-loop Control Systems**: These systems are sensitive to ...}

\end{tcolorbox}
\begin{tcolorbox}[mylistingstyle,
  title=ProjectManager, 
  coltitle=black, fonttitle=\bfseries, fontupper=\small,colbacktitle=blue!20, colframe=black, colback=white,
  left=1mm, right=1mm, top=1mm, bottom=1mm, boxsep=1mm]
\texttt{SafetyEngineer, now that we have identified the control-related root causes of excessive swinging and overheating, let's explore the safety implications of these issues. Please provide your insights on ...}

\end{tcolorbox}
\begin{tcolorbox}[mylistingstyle, colframe=black, colback=white,fonttitle=\bfseries,left=1mm, right=1mm, top=1mm, bottom=1mm, boxsep=1mm]
\centering
\texttt{...}

\end{tcolorbox}

\end{tcolorbox}

\caption{Example of messages in conversation between agents during Step 3.}
\label{fig:conversation}
\end{minipage}
\end{figure}

\paragraph{Step 3: Cause and Effect Chain Analysis (CECA)} Similar actions were taken by the agent while working on Cause and Effect Chain Analysis. The \texttt{Control Systems Engineer} was called to make the analysis and used a web search tool. Then \texttt{Safety Engineer} contributed to this step. The step ended with documentation on the \texttt{Project Manager}'s request. Part of the conversation of step 3 was shown in Figure \ref{fig:conversation}.

\paragraph{Step 4: Engineering Contradiction (EC) and Contradiction Matrix.} \texttt{Project Manager} asks only \texttt{TRIZSpecialist} agent to do this task. The agent first checks the list of TRIZ features from which it can choose. Based on that, it identifies contradictions.
The agent uses another TRIZ tool, a contradiction matrix, to find inventive principles for those contradictions. Next, it uses another tool that provides more details about contradictions, which will help find a solution.

\paragraph{Step 5: Physical Contradiction} This step is also solely performed by \texttt{TRIZ Specialist} by one answer only without any tool use. With this information, \texttt{Project Manager} ended this step with a documentation request.

\paragraph{Step 6: Solutions} Having all previous steps documented, \texttt{Project Manager} decided to begin the last step, which was finding solutions based on identified innovative principles. In this step, an agent who contributed first was \texttt{Control Systems Engineer}, who used a web search tool to enhance its findings. The following two agents who reviewed the proposed solutions were \texttt{Safety Engineer} and \texttt{Operations Specialist}. After solution propositions, \texttt{Project Manager} directed the end of collaborative work. This resulted in compiling all previously created documents into one final report.

\subsection{Assessment and Discussion}

In this subsection, we will assess the actions performed by the team. We will first evaluate its performance as a team and then compare it with solutions proposed by the case study's original research team. Two separate runs of TRIZ Agents were chosen for assessment, and their results were compiled.

\subsubsection{TRIZ Agents Actions}
Overall, all steps were executed, and in the end, the team proposed solutions to the problem. There were steps where the outcome was the result of interaction between multiple agents (steps: 2, 3, 6), and there were steps where only one agented contributed to the result (steps: 1, 4, 5) - not counting the manager who was only responsible for leading the discussion and not ideation. How each step was solved and who contributed to it was solely the decision of the \texttt{Project Manager}. It is desirable that multiple agents cooperate to solve each step because they may correct each other's outputs. This suggests adding more critique instructions to \texttt{Project Manager} agent's prompt for doubting the outputs of other agents. Another option would be replacing each worker agent with a team of agents performing one task. This would create a hierarchical teams architecture, a promising but more complex approach to agent orchestration.

Another observation is that the \texttt{TRIZ Specialist} rarely used the RAG tool to request more information about the TRIZ methodology. Although the agent's profiling prompt mentioned the RAG tool, it was not explicitly stated that it should be utilized at every generation, which may explain this behavior. \texttt{TRIZ Specialist} agent was intentionally provided with autonomy to decide whether to use external resources, implicitly allowing it to rely on its internal knowledge. Aim was to ensure that the agent follows the TRIZ methodology closely, ideally using the context of TRIZ literature. However, this autonomy caused the agent to underuse the RAG tool, as the agent often generated responses without any tool usage or with the usage of the web search tool. This outcome demonstrates one of the challenges of prompt engineering - balancing agent autonomy with consistent obedience to certain assumptions about the agent. To address this, future work will examine improving the prompts to require or encourage using RAG tools.

\subsubsection{Case Study Comparison}
There are similarities between solutions, as shown by comparing the result to the original case study's solution. This section will focus on comparing those two results.

\paragraph{Step 1: Defining Engineering System} Most of the Sub/System and System Components have been found. In some, the name may be slightly different, but the object is the same. Supersystems were not as well identified. Results are compared in Table \ref{tab:step1}.

\begin{table}[!ht]
\centering
\renewcommand{\arraystretch}{1.4} 
\setlength{\tabcolsep}{4pt} 
\begin{tabularx}{\columnwidth}{|>{\centering\arraybackslash}X|}
\hline
\textbf{Gantry Legs}, Wheel, \textbf{Railway}, \textbf{Trolley}, Trolley Frame, 
\textbf{Wire Rope}, Motor, \textbf{Hoist}, \textbf{Hook}, Railway Beam, \textbf{Bridge Girder} \\ \hline
Air particles/dust, \textbf{Workers}, Humidity, \textbf{Thermal} \\ \hline
\end{tabularx}
\caption{Elements of the engineering system in the case study: the upper row represents the system components, while the lower row represents the supersystems. Elements in bold were identified by TRIZ agents.}
\label{tab:step1} 
\end{table}

\paragraph{Step 2: Function Analysis} In the case study, this step was presented using a graph showing functions as connections between elements. TRIZ Agents listed the connections. Those connections labeled as useful connections that overlap are: \textit{Trolley to Hoist (Motor)}, \textit{Wire Rope to Hook}. Connection \textit{Hoist to Wire Rope} was identified as useful, but case study researchers label it as insufficient and harmful. On the other hand, the connection \textit{Hoist to Load} was identified as harmful, which is comparable. 
The connection \textit{Environmental Conditions to Load} was successfully identified, as the case study analysis highlights \textit{Dust}, \textit{Humidity}, and \textit{Thermal} factors. However, six useful connections present in the case study were not identified. Additionally, the harmful connection between \textit{Worker and System} was missed by the TRIZ Agents.

\paragraph{Step 3: Cause and Effect Chain Analysis (CECA)} case study identified two main root causes, which are \textit{the crane operator who does not follow standard operational procedure by lifting the excessive weight of load} and \textit{fast speed of controlling the movement of the cranes}. Those causes were identified by TRIZ Agents as \textit{Overloading and Rapid Movements}. The system identified five other root causes that are logical but do not occur in the case study.

\paragraph{Step 4: Engineering Contradiction (EC) and Contradiction Matrix.} TRIZ Agents identified two engineering contradicitons: \textit{Speed} vs. \textit{Stability} and \textit{Load Capacity} vs. \textit{Safety}. This does not exactly overlap with contradictions from a research paper. However, TRIZ agents were not far from their solution because they identified \textit{Stability} instead of \textit{Object generated harmful factors} and \textit{Load Capacity} instead of \textit{Weight of stationary object}, which seem close enough. Inventive principles were then found based on those contradictions using the Contradiction Matrix, which is a deterministic operation, which is why there is no point in comparing those.

\paragraph{Step 5: Physical Contradiction} Physical contradicitons were found exactly the same. Results are presented in Table \ref{tab:pc}. In the next step, TRIZ Agents derived only one, common with case study, inventive principle. This may be due to the generally scarce literature sources about physical contradictions identification compared to engineering contradictions.

\noindent
\begin{table}[!ht]
\centering
\renewcommand{\arraystretch}{1.5} 
\setlength{\tabcolsep}{6pt} 
\begin{tabular}{|>{\centering\arraybackslash}p{0.26\columnwidth}|p{0.60\columnwidth}|}
\hline
\textbf{Contradiction} & \textbf{Contradictory Needs} \\ \hline
Speed & 
The crane must move quickly to enhance productivity, but it must also move slowly to prevent load swinging. \\ \hline
Load Capacity & 
The crane must lift heavy loads to improve efficiency, but it must lift lighter loads to ensure safety and prevent overheating. \\ \hline
\end{tabular}
\caption{Physical Contradictions identified by TRIZ Agents, which overlap with case study.}
\label{tab:pc}
\end{table}

\paragraph{Step 6: Solutions} Theoretical solutions proposed were categorized based on the specializations of the agents contributing to the ideation process. The first solution is the application of \textit{Sliding Mode Control (SMC) with Antiswing Trajectory}, which matches the case study's first proposed solution. The solution proposed by case study researchers \textit{intelligent circuit breaker with essential sensors to monitor and regulate the current consumption} was not found by TRIZ Agents. During Step 6, the Electrical Engineering agent was not involved in collaboration, which may be the primary reason for this shortfall, as this agent could have contributed the necessary idea in the electrical domain. However, TRIZ Agents mention \textit{Thermal Management Enhancements}, which is in the area of the last case study solution \textit{smart ventilation system with features like self-cleaning filters and sealed bearings}.

\section{CONCLUSIONS}
\label{sec:conclusion}

\subsection{Main insights}

The study showed that it is possible to model group work on innovative problems using the TRIZ methodology with a multi-agent system based on LLMs. The outputs of the TRIZ Agents' steps were not identical to those in the case study in every detail, but each step had commonalities. The solutions proposed by TRIZ Agents, although different from the case study, seemed logical. Of course, such multi-agent systems are still far from achieving a complete and adequate ideation process, but they can quickly produce solutions or guide real-world researchers in a promising direction.
Another insight is that the outcome of the system strongly depends on agents' profiling prompts. Experimenting with many prompts eventually resulted in the team following the workflow.

\subsection{Limitations}
A substantial limitation of the proposed system is its lack of a feedback loop in the teams' actions. In theory, it is possible, but the profiling prompt of \texttt{Project Manager} does not mention doing it. The supervisor agent never directed returning to any step to refine results or try another approach. This is a severe drawback because every problem-solving process requires iterating over ideas, revisiting steps, and rethinking previous actions. The absence of such an essential component is due to the high complexity of implementing such a feature. For example, creating a loop in agentic systems carries a significant risk of causing never-ending conversations between agents. This limitation poses a promising research direction: making multi-agent systems even more similar to real-world teams. 

There are also limitations connected with LLMs properties like context window. In the discussed use case, six workflow steps were outlined. In the case of a workflow with more steps, the context window could be too small to store all documentation from previous steps. In fact, step documentation alone is a way to handle this problem because instead of a long list of messages, a summary is stored. This sparks a need for long-time agent memory, where information is available when necessary \cite{hatalis2023memory}.

\subsection{Future Work}
Studies have shown a promising and worth-exploring domain of multi-agent LLM systems. Examining how other agent orchestration architectures might perform would probably bring interesting insights. Additionally, only the TRIZ methodology is considered in the study. However, different methods, such as Lean Innovation \cite{sehested2010lean} and Design Thinking \cite{brown2008design}, are worth investigating as they may work better with LLMs.

A promising direction for further research is to compare how different prompt strategies function in a multi-agent system scenario. Refining prompts so that agents iterate over ideas, repeat, and rethink previous steps and actions could lead to substantial improvements in the system. Although this research did not focus on comparing prompts, it was surely a significant part of the implementation part because prompts drastically influenced the system's actions. Applying an optimal prompt strategy might enhance the results. 

Another interesting field to investigate is cognitive architectures in multi-agent systems. Cognitive architectures are augmented LLMs agents with the internal flow or external sources for tasks with grounding or reasoning \cite{sumers2024cognitivearchitectureslanguageagents}. Building cognitive agents could enable the system to learn problem descriptions. There are already interesting studies about using those concepts for enhancing decision-making \cite{wucognitive} and problem-solving \cite{sun2024can}.

\bibliographystyle{apalike}
{\small
\bibliography{references}}

\end{document}